\documentclass{article}

\usepackage{amsmath,amssymb,amsthm,color}
\usepackage{graphicx}
\usepackage{epsfig}
\usepackage{tabularx, multirow, booktabs}

\usepackage{times}

\usepackage{amssymb,amsmath}
\usepackage{epsfig,graphicx}
\usepackage{verbatim}
\usepackage{named}
\usepackage{amsfonts}

\long\def\comment#1{}

\newcommand{\D}{\:\: |\:\:}

\newtheorem{theorem}{Theorem}

\newtheorem{examp}{Example}
\newtheorem{srss}{Structured Rule System}

\newenvironment{example}{\begin{examp}\rm}{\end{examp}}

\begin{document}

\title{A Set Theoretic Approach for Knowledge Representation: the Representation Part}

\author{Yi Zhou \\ Artificial Intelligence Research Group \\ Western Sydney
University, NSW Australia}

\maketitle

\begin{abstract}
In this paper, we propose a set theoretic approach for knowledge
representation. While the syntax of an application domain is
captured by set theoretic constructs including individuals, concepts
and operators, knowledge is formalized by equality assertions. We
first present a primitive form that uses minimal assumed knowledge
and constructs. Then, assuming naive set theory, we extend it by
definitions, which are special kinds of knowledge. Interestingly, we
show that the primitive form is expressive enough to define logic
operators, not only propositional connectives but also quantifiers.
\end{abstract}

\section{Introduction}\label{sec-intro}

Knowledge representation and reasoning (KR) is one of the central
focuses of Artificial Intelligence (AI)
\cite{Baral03,Brachman04,Harmelen08}. KR intends to syntactically
formalize information in an application domain as knowledge. Then,
complex problems in the domain can be solved by reasoning about the
knowledge. KR is not only of its own interests but also highly
influential to many other subfields in AI, including expert systems,
multi-agent systems, planning, uncertainty, cognitive robotics and
semantic Web \cite{Brachman04,Harmelen08}.

Nevertheless, KR has encountered huge difficulties. One issue is
that there are too many features and building blocks to be
considered \cite{Harmelen08}, for instance, propositions, variables,
connectives, rules, actions, common sense, time/space, uncertainty,
mental states and so on. In fact, KR has made huge successes on
formalizing these building blocks separately. However, combing them
together, even several of them, seems an extremely difficult task.
On the other side, many application domains, e.g. robotics, need
multiple building blocks at the same time.

\comment{ Another issue is that KR approaches are sometimes
considered not very user-friendly, that is, not easy to be
understood and to be used by knowledge engineers. This point might
be controversial. We use propositional logic as a case study here.
Propositional logic seems simple from a KR researcher's point of
view, which is simply nested applications of propositions and basic
logic connectives. Nevertheless, from our own experiences as a
teacher, many undergraduate and graduate students --- the future
knowledge engineers, have difficulties in representing knowledge by
using propositional logic, especially when multiple nested operators
are needed. Yet propositional logic is a foundation of many other KR
formalisms.}

Another critical related issue is about the balance between
expressiveness and efficiency. It is widely believe that the more expressive, the less efficient,
and vice versa \cite{LevesqueB87}. However, in many
application domains, e.g., robotics, we need both. Yet this is a
very difficult task for KR formalisms. Consider propositional logic,
a fundamental KR formalism that only takes propositions and natural
propositional connectives into account. The inference problem is
coNP complete, which means that, most likely, it takes exponential
time in worst case.

\comment{
Although many arguments
have been made to ignore worst case complexity \cite{} and many
advanced techniques have been developed in SAT/ASP solving \cite{},
the balance between expressiveness and efficiency is still a critical barrier for KR.
}

Against this backdrop, we argue that KR needs a simple, extensible,
expressive and efficient approach. ``Extensible" means that this
approach should allow a current system in it to be
easily extended with more building blocks. ``Expressive" means that
this approach should be able to represent different types of
knowledge for a given set of building blocks.``Efficient" means that
this approach can efficiently reason about and derive new knowledge
in order to solve problems. Nevertheless, ``simple" is an ambiguous
term, which can be further elaborated into three aspects: primitive
--- using minimal primitive constructs, succinct --
able to represent knowledge in various application domains with relatively small
knowledge bases and user-friendly --- easy to be understood and used by knowledge engineers.

\comment{
The above goal seems too good to be true. For instance, it is widely
believed that expressiveness and efficiency cannot be achieved at
the same time. The more expressive, the less efficient. Likewise,
the more efficient, the less expressive. One has to make a good
balance between them \cite{}. Also, simplicity and
expressiveness/extensibility seem hardly to live happily together.

Nevertheless, we believe that this is not a mission impossible.}

Towards this goal, we propose a set theoretic knowledge
representation approach for syntactically representing knowledge in
application domains. While the syntax of an application domain is
captured by a set of individuals, concepts and operators, knowledge
is simply formalized by equality assertions of the form $a=b$, where
$a$ and $b$ are either atomic individuals or compound individuals.
Semantically, individuals, concepts and operators are interpreted as
elements, sets and functions respectively in naive set theory and
knowledge of the form $a=b$ means that the two individuals $a$ and
$b$ are referring to the same element.

We separate representation and reasoning. This paper is mainly
concerned with the basic ideas and the representation part, and we
leave the reasoning part to another paper. In this paper, we first
present the primitive form that uses minimal assumed knowledge and
primitive constructs. Then, assuming naive set theory, we extend it
with more building blocks by definitions that use assertions to
define new syntactic objects, including individuals, concepts and
operators. Once these new objects are defined, they can be used as a
basis to define more. As an example, we show that we can define
multi-assertions by using Cartesian product, and nested assertions
by using multi-assertions. Interestingly, we show
that this method, i.e., extending the primitive form by definitions
based on native set theory, is powerful enough to syntactically
capture logic operators, including both propositional connectives
and quantifiers.

\section{The Primitive Form}\label{sec-pf}

In this section, we present the primitive form of our set theoretic
knowledge representation approach. As the goal is to
syntactically represent knowledge in application domains, there are two essential tasks, i.e., how to capture the
syntax of the domain and how to represent knowledge in it.

\comment{
Roughly
speaking, we use set constructs, including individuals, concepts and
operators for the former task while we use equality assertions for
the latter.
}

We assume naive set theory \cite{Halmos60}, including the basic
concepts such as elements, sets and functions, Cartesian product,
the built-in relationships $\in$ and $\subseteq$, the built-in
operators $\cup$, $\cap$ and $\setminus$, the Boolean set
$\{\top,\bot\}$ and the set $\mathbb{N}$ of natural numbers,
cardinality and set specifications.

\subsection{Capturing the syntax}

Given an application domain, a {\em syntactic structure} ({\em
structure} for short if clear from the context) of the domain is a
triple $\langle \mathcal{I}, \mathcal{C}, \mathcal{O} \rangle$,
where $\mathcal{I}$ is a collection of {\em individuals},
representing objects in the domain, $\mathcal{C}\subseteq
2^\mathcal{I}$ a collection of {\em concepts}, representing groups
of objects sharing something in common and $\mathcal{O}$ a
collection of {\em operators}, representing relationships and
connections among individuals and concepts.

Concepts and operators can be considered as individuals as well.
If needed, we can have concepts of concepts, concepts of
operators, concepts of concepts of operators and so on.

An operator could be multi-ary, that is, it maps a tuple of
individuals into a single individual.\footnote{Note that in naive
set theory, a tuple of sets is a Cartesian product of some sets,
which itself is a set as well. Therefore, multi-ary operators can
essentially be viewed as single-ary.} Each multi-ary operator $O$ is
associated with a {\em domain} of the form $(C_1, \dots, C_n)$,
representing all possible values that the operator $O$ can operate
on, where $C_i, 1 \le i \le n$, is a concept. We call $n$ the {\em
arity} of $O$. For a tuple $(a_1,\dots,a_n)$ matching the domain of
an operator $O$, i.e., $a_i \in C_i, 1 \le i \le n$, $O$ maps
$(a_1,\dots,a_n)$ into an individual, denoted by $O(a_1,\dots,a_n)$.
We also use $O(C_1,\dots,C_n)$ to denote the set $\{O(a_1,\dots,a_n)
\D a_i \in C_i\}$, called the {\em range} of the operator $O$.

Operators are similar to functions in first-order logic but
differs in two essential ways. First, operators are many-sorted as
$C_1, \dots, C_n$ could be different concepts. More importantly,
$C_1, \dots, C_n$ could be high-order constructs, e.g., assertions,
concepts of concepts, concepts of operators and so on.

For instance, consider the arithmetic domain, in which $0$, $1$,
$2$, etc., are individuals; the set $\mathbb{N}$ of natural numbers
is a concept; the successor operator $Succ$ and the add operators
$Add$ are operators.

For convenience, if $O$ is unary, we sometimes use $a.O$ ($C.O$) to
denote $O(a)$ ($O(C)$), where $a \in \mathcal{I}$ and $C \in
\mathcal{C}$. If $O$ is binary, we sometimes use $a\:O\:b$ ($A\: O
\: B$) to denote $O(a,b)$ ($O(A,B)$), where $a, b \in \mathcal{I}$
and $A,B \in \mathcal{C}$. If the range of an operator $O$ is
Boolean, we sometimes use $O(a_1,\dots,a_n)$ to denote
$O(a_1,\dots,a_n)=\top$.

\comment{ We use pre-given, established syntactic objects
(individuals, concepts and operators) to define new ones. The former
is said to be {\em prior} while the latter is said to be {\em
posterior}. Prior objects are assumed knowledge while posterior
objects are knowledge to be studied. Prior and posterior objects are
relative with respect to a particular application domain. For
instance, when we use the successor operator to define the add
operator, the former is prior whilst the latter is posterior.
Nevertheless, when we use the add operator to define something else,
it becomes prior. }

\subsection{Representing knowledge}

Let $\langle \mathcal{I}, \mathcal{C}, \mathcal{O} \rangle$ be a
syntactic structure. A {\em term} is an individual, either an atomic
individual $a\in \mathcal{I}$ or the result $O(a_1,\dots,a_n)$ of an
operator $O$ operating on some individuals $a_1,\dots,a_n$. We also
call the latter {\em compound individuals}.

\comment{ A term is said to be {\em prior} if all individuals and
operators in it are prior. Otherwise, it is said to be {\em
posterior}.}

An {\em assertion} is of the form
\begin{equation}\label{form-assertion}
a=b,
\end{equation}
where $a$ and $b$ are two terms. Intuitively, an assertion of the
form (\ref{form-assertion}) is a piece of knowledge in the
application domain, claiming that the left and right side are
referring to the same objects. Here, $=$ is the the built-in
equality relation in naive set theory. Hence, $a=b$ can be
understood in alternative way that $=(a,b)$ is true. A {\em
knowledge base} is a set of assertions. Terms and assertions can be
considered as individuals as well.

For instance, in arithmetic, $0=Succ(1)$ and $2+3=5$ are two typical
assertions.

\comment{Assertion consistent, deterministic}

Similar to concepts that group individuals, we use schemas to group
terms and assertions. A {\em schema term} is either an atomic
concept $C\in \mathcal{C}$ or the collection of results
$O(C_1,\dots, C_n)$. Essentially, a schema term represents a set of
terms, in which every concept is grounded by a corresponding
individual. Then, a {\em schema assertion} is of the same form as
form (\ref{form-assertion}) except that terms can be replaced by
schema terms. Similarly, a schema assertion represents a set of
assertions.

\comment{
We say that a schema term $T$
{\em mentions} a set $\{C_1,\dots,C_n\}$ of concepts if
$C_1,\dots,C_n$ occur in $T$, and {\em only mentions} if
$\{C_1,\dots,C_n\}$ contains the set of all concepts mentioned in
$T$.
}

\comment{
Similarly, we say that a schema assertion {\em only mentions} a set
$\{C_1,\dots,C_n\}$ of concepts if $\{C_1,\dots,C_n\}$ contains the
set of all concepts mentioned in $A$.
}

Note that it could be the case that two or more different
individuals are referring to the same concept $C$ in schema terms
and assertions. In this case, we need to use different {\em copies}
of $C$, denoted by $C^1,C^2,\dots$, to distinguish among them. For
instance, all assertions $x=y$, where $x$ and $y$ are numbers, are
captured by the schema assertion $\mathbb{N}^1=\mathbb{N}^2$. On the
other side, in a schema, the same copy of a concept $C$ can only
refer to the same individual. For instance, $\mathbb{N}=\mathbb{N}$
is the set of all assertions of the form $x=x$, where $x \in
\mathbb{N}$.

\subsection{The semantics}

We introduce a set theoretic semantics to define the meanings of
syntactic structures and knowledge. An {\em interpretation} is a
pair $\langle \Delta, .^{I} \rangle$, where $\Delta$ is a domain of
elements that admits naive set theory and $.^{I}$ is a mapping
function that maps individuals into domain elements in $\Delta$,
concepts into sets in $\Delta$ and operators into functions in
$\Delta$. The mapping functions $.^{I}$ can be generalized into
mapping from terms to elements.

Let $I$ be an interpretation and $a=b$ an assertion. We say that $I$
is a {\em model} of $a=b$, denoted by $I \models a=b$ iff
$.^{I}(a)=.^{I}(b)$, also written $a^I=b^I$. Let $KB$ be a knowledge
base. We say that $I$ is a model of $KB$, denoted by $I \models KB$,
iff $I$ is a model of every assertion in $KB$. We say that an
assertion $A$ is a {\em property} of $KB$, denoted by $KB \models
A$, iff for all interpretations $I$ such that $I \models KB$, we
have $I \models A$. In particular, we say that an assertion $A$ is a
{\em tautology} iff it is modeled by all interpretations.

Since we assume naive set theory, we directly borrow some set
theoretic constructs on individuals, concepts and operators. For
instance, we can use $\cup (C_1, C_2)$ (also written as $C_1 \cup
C_2$) to denote a new concept that unions two concepts $C_1$ and
$C_2$. Applying this to assertions, we can see that assertions of
the form (\ref{form-assertion}) can indeed represent many important
features in knowledge representation. For instance, the {\em
membership assertion}, stating that an individual $a$ is an instance
of a concept $C$ is the following assertion $\in(a,C)=\top$ (also
written as $a \in C$). The {\em containment assertion}, stating that
a concept $C_1$ is contained by another concept $C_2$, is the
following assertion $\subseteq(C_1,C_2)=\top$ (also written as $C_1
\subseteq C_2$). The {\em range declaration}, stating that the range
of an operator $O$ operating on some concept $C_1$ equals to another
concept $C_2$ is the following assertion $O(C_1)=C_2$.

\section{Definitions for Extensibility}

The primitive form is a foundation that uses minimal assumed
knowledge and primitive constructs. Nevertheless, sometimes it is
not convenient to use it for formalizing an application domain,
e.g., to represent logic expressions. Hence, we extend it with more
building blocks. As discussed in the introduction section,
extensibility is a critical issue for KR approaches.

For this purpose, we introduce {\em definitions} in our approach.
Definitions use (schema) assertions to define new syntactic objects
(individuals, concepts and operators) based on existing ones. Note
that definitions are nothing extra but special kinds of knowledge.

\subsection{Defining individuals, operators and concepts}

We start with defining new individuals. An individual definition is
a special kind of assertion of the form
\begin{equation}\label{form-individual-definition}
a=t,
\end{equation}
where $a$ is an atomic individual and $t$ is a term. Here, $a$ is the
individual to be defined. This assertion claims that the left side
$a$ is defined as the right side $t$. For instance, $0=\emptyset$
means that the individual $0$ is defined as the empty set.

Defining new operators is similar to defining new individuals except
that we use schema assertions for this purpose. Let $O$ be an
operator to be defined and $(C_1,\dots,C_n)$ its domain. An operator
definition is a schema assertion of the form
\begin{equation}\label{form-operator-definition}
O(C_1,\dots,C_n)=T,
\end{equation}
where $T$ is a schema term that mentions concepts only from
$C_1,\dots,C_n$. It could be the case that $T$ only mentions some of
$C_1, \dots, C_n$. Note that if $C_1,\dots,C_n$ refer to the same
concept, we need to use different copies respectively.

Since a schema assertion represents a set of assertions,
essentially, an operator definition of the form
(\ref{form-operator-definition}) defines the operator $O$ by
defining the value of $O(a_1,\dots,a_n)$ one-by-one, where $a_i \in
C_i, 1 \le i \le n$. Sometimes we also define operators in this way.
For instance, for defining the successor operator $Succ$, we can use
the schema assertion $Succ(\mathbb{N})=\{\mathbb{N},
\{\mathbb{N}\}\}$. This is equivalent to an alternative definition
stating that, for every natural number $n$, the successor of $n$, is
defined as $\{n,\{n\}\}$, i.e., $Succ(n)=\{n,\{n\}\}$. For instance,
$Succ(0)$ is defined as $\{\emptyset, \{\emptyset\}\}$.

Defining new concepts is different. As concepts are essentially
sets, they are defined through set theoretic constructions. We
directly borrow set theory notations to define concepts as follows:

\noindent {\bf Enumeration} Let $a_1,\dots,a_n$ be $n$ individuals.
Then, the collection $\{a_1,\dots, a_n\}$ is a concept, written as
\begin{equation}
C=\{a_1,\dots, a_n\}.
\end{equation}
For instance, we can define the concept $Digits$ by
$Digits=\{0,1,2,3,4,5,6,7,8,9\}$.

\noindent {\bf Operation} Let $C_1$ and $C_2$ be two concepts. Then,
$C_1 \cup C_2$ (the union of $C_1$ and $C_2$), $C_1 \cap C_2$ (the
intersection of $C_1$ and $C_2$), $C_1 \setminus C_2$ (the
difference of $C_1$ and $C_2$), $C_1 \times C_2$ (the Cartesian
product of $C_1$ and $C_2$), $2^{C_1} $ (the power set of $C_1$) are
concepts. Operation can be written by assertions as well. For
instance, the following assertion
\begin{equation}
C=C_1 \cup C_2
\end{equation}
states that the concept $C$ is defined as the union of $C_1$ and
$C_2$. As an example, one can define the concept $Man$ by
$Man=Human\cap Male$.

\noindent {\bf Comprehension} Let $C$ be a concept and $A(C)$ a
schema assertion that only mentions concept $C$. Then, individuals
in $C$ satisfying $A$, denoted by $\{x\in C | A(x)\}$ (or simply
$C|A(C)$), form a concept, written as
\begin{equation}
C'=C|A(C).
\end{equation}
For instance, we can define the concept $Male$ by $Male =\{Animal \D
Sex(Animal)=male\}$, meaning that $Male$ consists of all animals
whose sex are male.

\noindent {\bf Replacement} Let $O$ be an operator and $C$ a concept
on which $O$ is well defined. Then, the individuals mapped from $C$
by $O$, denoted by $\{O(x) \D  x \in C\}$ (or simply $O(C)$), form a
concept, written as
\begin{equation}
C'=O(C).
\end{equation}
For instance, we can define the concept $Parents$ by
$Parents=ParentOf(Human)$, meaning that it consists of all
individuals who is a $ParentOf$ some human.

Definitions can be incremental. We may define some syntactic objects
first. Once defined, they can be used to define more. One can always
continue with this incremental process to extend the current system.
For instance, in arithmetic, we define the successor operator first.
Once defined, it can be used to define the add operator, which is
further served as a basis to define more and more useful syntactic
objects.

For clarity, we use the symbol ``$::=$" to replace ``$=$" for definitions. We force
uniqueness of definitions. That is, each syntactic object can only
be defined at most once.

Another critical issue is about recursiveness. Clearly, a definition
such as $a::=a+1$ is invalid and meaningless. Hence, we need to
restrict our definitions. However, sometimes we do use recursion to
define concepts. For instance, in arithmetic, natural numbers are
define as $\mathbb{N} ::= \{0\} \cup Succ(\mathbb{N})$, meaning that
if $n$ is a natural number, then the successor of $n$, i.e.
$Succ(n)$, is also a natural number.

We require that recursion can only be used in replacement definition
of concepts. When a recursive replacement is used, we interpret it
as an infinite process. At the beginning, all concepts contain and
only contain those individuals defined by non-replacement
definitions. Then, we apply the replacement definitions to obtain
new versions of concepts. This finishes the first step. We continue
with the process. At each step, we first use those non-replacement
definitions to expand the concepts. Then, again, we apply the
replacement definitions to obtain new versions of concepts. This
could be an infinite process. For instance, consider the definition
of natural numbers. Initially, we have $\{0\}$. Then, applying the
replacement definition, we expand it to $\{0,Succ(0)\}$. We continue
with this process to obtain the infinite set of natural numbers
$\{0,Succ(0),Succ(Succ(0)),\dots\}$.

We require that all other definitions are non-recursive. Formally,
the {\em definition dependency graph} over a set of definitions
(without replacements) is a directed graph $<V,E>$, where $V$
consists of all syntactic objects appeared in these definitions and
$E$ is the set of all pairs $(a,b)$ such that there exists a
definition whose left side is $a$ and whose right side mentions
$b$.\footnote{For operator definitions, we ignore the concepts that
are arguments in the operator and in the schema term since they are
essentially grounded into individuals. } A set of definitions is
said to be {\em non-recursive} if its corresponding definition
dependency graph is acyclic.

In fact, one can observe that the Backus-Naur form (BNF), widely
used in computer science to define syntax, can be considered as a
special case of our concept definitions over strings. More
precisely, BNF only uses three features, enumeration (of a single
element), union operation and recursive replacement by using the
pre-assumed concatenation operator. Comprehension and other set
operations are not used.

\subsection{Multi-assertions}\label{sec-multi}

As a case study of extending the primitive form by definitions, we
extend assertions of the form (\ref{form-assertion}) into
multi-assertions.

\comment{
Recall the notion of Cartesian product. Let $S_1,S_2, \dots, S_n$ be
$n$ sets. The Cartesian product of $S_1,S_2, \dots, S_n$, denoted by
$S_1 \times S_2 \times \dots \times S_n$, is the set
\[
\{(a_1,a_2,\dots,a_n) \D a_i \in S_i, 1 \le i \le n\}.
\]

In fact, we have already used Cartesian product to define multi-ary
operators in Section \ref{sec-pf}. Multi-ary operators can be
defined alternatively in two steps as follows. First, we start with
single-ary operators, which are functions mapping from a single
individual to another. Then, multi-ary operators can be defined by
allowing domains of single-ary operators to be Cartesian products of
some concepts. This way treats single-ary operators as the primitive
constructs while multi-ary operators as an extension. In this sense,
it is even more primitive. Nevertheless, for clarity and
convenience, in Section \ref{sec-pf}, we directly allow operators to
be multi-ary.
}

Given a number $n$, we define a new operator $M_n$ for
multi-assertions with arity $n$ by the following schema assertion:
\small
\begin{equation*}
M_n(C_1=D_1,\dots,C_n=D_n)::=(C_1,\dots,C_n)=(D_1,\dots,D_n),
\end{equation*}
\normalsize where $C_i, D_i, 1 \le i \le n$, are concepts of terms.
This assertion states that for $n$ assertions, $Assertion_i, 1\le i
\le n$, of the form (\ref{form-assertion}), namely $a_i=b_i$,
$M_n(a_1=b_1,\dots,a_n=b_n)$ is $ (a_1,\dots, a_n) = (b_1,
\dots,b_n)$. Hence, $M_n(a_1=b_1,\dots,a_n=b_n)$ holds if and only
if for all $i, 1 \le i \le n$, $a_i=b_i$, that is, $Assertion_i$
holds. In this sense, this single assertion can be used to represent
$n$ assertions $Assertion_i, 1\le i \le n$.

Then, we define the concept of multi-assertions as follows:
\[
Multi-Assertion ::= \bigcup_{1 \le i \le \infty}
M_i(\mathcal{A}^1,\dots,\mathcal{A}^i),
\]
where $\mathcal{A}^1,\dots,\mathcal{A}^i$ are $i$ copies of standard
assertions. For convenience, we use $ Assertion_1, \dots,
Assertion_n$ to denote a $n$-ary multi-assertion.

\comment{
This operation definition claims that the set of
multi-assertions are defined as the union of all $n$-ary
multi-assertions.
}

Note that multiple assertions are just
syntactic sugar of the primitive form as they can be defined in the primitive form by using
ordered pairs and Cartesian products. In this sense, they do not increase the expressive power
of the primitive form. Nevertheless, using them can make the
representation task more convenient in some cases. Multi-assertions are not only of interests themselves. Once
defined, they can be used to define more syntactic building
blocks. Note that finite knowledge bases are finite sets of assertions, i.e., multi-assertions, which can
essentially be viewed as single assertions.

\comment{
\subsection{General assertions}

As an example, we use multi-assertions to define general assertions.
Note that in the assertion form (\ref{form-assertion}), the right
side has to be a prior term. This can be lifted into general
assertions of the form
\[
a=b,
\]
where both $a$ and $b$ can be posterior terms. In fact, a general
assertion of the above form can be re-written as a multi-assertion
\[
a=i, b=i,
\]
where $i$ is a newly introduced prior individual. Notice that if a
general assertion $a=b$ is used, $a$ and $b$ must be assumed to
refer individuals from the same prior concept, e.g., natural
numbers. }

\subsection{Nested terms and assertions}

We continue with our extensions for the primitive form by
introducing nested terms and nested assertions. Note that terms
defined in Section \ref{sec-pf} cannot be nested in the sense that
individuals used inside an operator must be atomic. This can be
generalized to nested terms, where operators can use compound
individuals inside.

Nested terms are defined by the following definition:
\begin{eqnarray*}
Nested-Term & ::= & Term \cup N-Term \\
N-Term & ::= & Op(Nested-Term),
\end{eqnarray*}
where $Term$ is the concept of standard term defined in Section
\ref{sec-pf}, $Op$ is an arbitrary operator and $Op(Nest-Term)$ is a
replacement definition such that individuals in $Nest-Term$ are in
the domain of $Op$.

The above definition is a recursive definition. In fact, it can be simplified as
\[
Nested-Term::=Term\cup Op(Nested-Term).
\]
However, this definition itself uses nested terms as well since
$Op(Nested-Term)$ is not atomic. Hence, before formally defining the
meaning of nested terms, we use the former.

Nevertheless, from this example, we can see how to interpret nested
terms. That is, whenever a nested term is used, we introduce a new
atomic individual to replace it, and claim that this atomic
individual defines the nested term. To formalize this idea, we also
need nested assertions,  in which terms used on both sides of the
assertion can be nested.
\[
Nested-Assertion::=Nested-Term = Nested-Term.
\]

As mentioned, nested assertions can be represented by non-nested
multi-assertions by introducing new individuals. Whenever a result
of nested term is used, we introduce a new individual to replace it
and claim that this new individual is defined as the nested term.
That is, for every nested term
$Op(a_1,\dots,Op'(b_1,\dots,b_m),\dots,a_m)$ occurred in a nested
assertion, we introduce a new atomic individual $a'$; replace the
above term with $Op(a_1,\dots,a',\dots,a_m)$ and add a new assertion
$a'=Op'(b_1,\dots,b_m)$. F(or instance, the nested assertion
$Op(a,Op(b,Op'(c)))=Op'(d)$ is defined as $Op(a,x)=Op'(d),
x=Op(b,y), y=Op'(c)$, where $x$ and $y$ are new individuals. In this
sense, nested assertion is essentially a multi-assertion, which can
be represented as a single assertion. Therefore, nested assertion is
a syntactic sugar of the primitive form as well.

Using nested assertions can simplify the representation task.
However, one cannot overuse nested assertions since, essentially,
every use of a nested term introduces a new individual. For
instance, one can easily get lost with a nested assertion like
$Op(a,Op(b,Op'(c)))=Op'(d)$.

\section{Logic Operators over Assertions}

\comment{
In the previous section, we showed how to use definitions,
i.e., special kinds of assertions, to define new syntactic objects
including individuals, operators and concepts. For instance, we can
define multi-assertions by using Cartesian product, and then nested
assertions by using multi-assertions. This shows the incremental
extensibility of our approach. Notice that both multi-assertions and
nested assertions do not bring new expressive power to the
approach as they are advanced uses of the primitive form based on naive set theory.
}

In this section, we continue to extend the primitive form with logic
operators over assertions. Interestingly, we can define not only
propositional connectives but also quantifiers based on naive set
theory. This, on one side, provides another case study how to extend
the primitive form, and on the other side, shows that, assuming
naive set theory, the primitive form is expressive enough to capture
logic.

\subsection{Propositional operators over assertions}

We start with the propositional case. Let $\mathcal{A}$ be the
concept of nested assertions. We introduce a number of operators
over $\mathcal{A}$, including $\neg(\mathcal{A})$ (for {\em
negation}), $\land(\mathcal{A}^1,\mathcal{A}^2)$ (for {\em
conjunction}), $\lor(\mathcal{A}^1,\mathcal{A}^2)$ (for {\em
disjunction}), $\to (\mathcal{A}^1,\mathcal{A}^2)$ (for {\em
implication}) and $\equiv(\mathcal{A}^1,\mathcal{A}^2)$ (for {\em
equivalence}).

There could be different ways to define those operators, depending on which operators
are defined directly and which are defined based on the previous ones. Here, we directly define
negation, conjunction, disjunction and implication and indirectly
define equivalence.

Let $a=a'$ and $b=b'$ be two (nested) assertions. The propositional
connectives are defined as follows:
\small
\begin{align*}
\neg(a=a')& ::= && \hspace{-.1in} \{a\}\cap\{a'\}=\emptyset \\
\land(a=a',b=b') & ::= && \hspace{-.1in} (\{a\}\cap\{a'\}) \cup
(\{b\}\cap\{b'\})=\{a,a',b,b'\} \\
\lor(a=a',b=b') & ::= && \hspace{-.1in} (\{a\}\cap\{a'\}) \cup
(\{b\}\cap\{b'\}) \ne \emptyset  \\
\to (a=a',b=b') & ::= && \hspace{-.1in} (\{a,a'\} \setminus
\{a\}\cap\{a'\})\cup (\{b\}\cap\{b'\}) \ne \emptyset \\
\equiv(a=a',b=b') & ::= && \hspace{-.1in}\land (\to (a=a',b=b'), \to
(b=b',a=a')).
\end{align*}
\normalsize We also use $a\ne a'$ to denote $\neg(a=a')$. One can
observe that the ranges of all logic operators are nested
assertions. Hence, similar to multi- and nested assertions,
propositional logic operators are syntactic sugar as well.

Now we consider some properties. For instance, according to the
definitions, De-Morgan's laws are tautologies.

\begin{theorem}
Let $A_1$ and $A_2$ be two (nested) assertions. Then, for all
interpretations $I$,
\[
I \models \neg(A_1 \lor A_2) \equiv \neg(A_1) \land \neg(A_2).
\]
\[
I \models \neg(A_1 \land A_2) \equiv \neg(A_1) \lor \neg(A_2).
\]
\end{theorem}

Also, the relationship between implication and
disjunction, i.e., $A_1 \to  A_2 \equiv \neg A_1 \lor
A_2$, is a tautology as well.

In fact, all tautologies in propositional logic are tautologies
under our context, i.e., modeled by all interpretations, and vice
versa. This, actually follows from the following theorem, stating
that the syntactic definitions above defines the semantics of logic
operators.
\begin{theorem}
Let $A_1$ and $A_2$ be two nested assertions. Then, for all
interpretations $I$,
\begin{itemize}
\item $I \models \neg(A_1)$ iff $I$ is not a model of $A_1$.
\item $I \models \land(A_1, A_2)$ iff $I$ is a model of both $A_1$
and $A_2$.
\item $I \models \lor(A_1,A_2)$ iff $I$ is a model of either $A_1$
or $A_2$.
\item $I \models \to (A_1,A_2)$ iff $I$ is a model of  $A_1$
implies that $I$ is a model of $A_2$.
\item $I \models \equiv(A_1,A_2)$ iff $I$ is a model of $A_1$ if and
only if $I$ is a model of $A_2$.
\end{itemize}
\end{theorem}

\comment{
\begin{theorem}
Tautologies in propositional logic is a tautology in our sense.
\end{theorem}
}

\comment{
We can define {\em propositional assertions} $\mathcal{PA}$ as
follows:
\[
\mathcal{PA} ::= \mathcal{A} \cup \neg(\mathcal{PA}) \cup
\mathcal{PA} \land \mathcal{PA} \cup \mathcal{PA} \lor \mathcal{PA}
\cup \mathcal{PA} \to  \mathcal{PA} \cup \mathcal{PA} \equiv
\mathcal{PA}.
\]
}

\subsection{Quantifiers over assertions}

Now we consider quantifiers, including $\forall$ (for the {\em
universal} quantifier) and $\exists$ (for the {\em existential}
quantifier). The domain of quantifiers is a pair $(C,A(C))$, where
$C$ is a concept and $A(C)$ is a schema assertion that only mentions
$C$.

The quantifiers are defines as follows:
\begin{eqnarray*}
\forall(C,A(C)) & ::= & C|A(C)=C\\
\exists(C,A(C)) & ::= & C|A(C) \ne \emptyset
\end{eqnarray*}
Intuitively, $\forall(C,A(C))$ is true iff those individuals $x$ in
$C$ such that $A(x)$ holds equals to the concept $C$ itself, that
is, for all individuals $x$ in $C$, $A(x)$ holds; $\exists(C,A(C))$
is true iff those individuals $x$ in $C$ such that $A(x)$ holds does
not equal to the empty set, that is, there exists at least one
individual $x$ in $C$ such that $A(x)$ holds. We can see that the
ranges of quantifiers are nested assertions as well. Thus,
quantifiers are also syntactic sugar of the primitive form.

Similarly, the syntactic definitions of quantifiers based on naive
set theory capture their semantics.
\begin{theorem}
Let $C$ be a concept and $A(C)$ a schema assertion that only
mentions $C$. For all interpretations $I$,
\begin{itemize}
\item $I \models \forall(C,A(C))$ iff for all individuals $a$ in $C$, $I \models A(a)$.
\item $I \models \exists(C,A(C))$ iff there exists at least one individual $a$ in $C$ such that $I \models A(a)$.
\end{itemize}
\end{theorem}

As a consequence,  one can prove some properties
about quantifiers. For instance, the universal quantifiers and
the existential quantifiers are dual under negation.

\comment{
\begin{theorem}
For all interpretations $I$, concept $C$ and assertion $A(C)$ that
only mentions $C$, $I \models \forall(C,A(C)) \equiv \neg \exists
(C,\neg A(C))$.
\end{theorem}
}

Note that quantifiers defined here are ranging from an arbitrary
concept $C$. If $C$ is a concept of all atomic individuals and all
quantifiers range from the same concept $C$, then these quantifiers
are first-order. Nevertheless, the concepts could be different. In
this case, we have many-sorted first-order logic. Moreover, $C$
could be complex concepts, e.g., a concept of all possible concepts. In
this case, we have monadic second-order logic. Yet $C$ could be many
more, e.g., a concept of assertions, a concept of concepts of terms
etc. In this sense, the quantifiers become high-order.

\comment{
We can define quantified assertions $\mathcal{QA}$ as follows:
\[
\mathcal{QA} ::= \mathcal{PA} \cup
\forall(\mathcal{C},\mathcal{PA}(\mathcal{C})) \cup
\exists(\mathcal{C},\mathcal{PA}(\mathcal{C})).
\]
}

\section{Conclusions, Discussions and Related Work}

In this paper, we have proposed a set theoretic approach to
syntactically represent knowledge in application domains. The syntax
of a domain is captured by individuals (i.e., objects in the
domain), concepts (i.e., groups of objects sharing something in
common) and operators (i.e., connections and relationships among
objects). From a set theory point of view, individuals, concepts and
operators are interpreted as elements, sets and functions
respectively. In the primitive form, knowledge in the domain is
simply captured by equality assertions of the form $a=b$, where $a$
and $b$ are terms.

We have shown how to extend a system by definitions, which are
special kinds of knowledge used to define new individuals, concepts
and operators. For instance, we have extended the primitive form with
multi-assertions and nested assertions, which are just syntactic
sugar of the primitive form as they can be expressed in it.
Extensibility is a critical issue for KR.  A KR approach should be able to define new
syntactic objects based on exiting ones. Once
defined, these objects can be further used to define more.

Interestingly, we have shown that logic operators, not only
propositional connectives but also quantifiers, can be defined in
the primitive form based on naive set theory. This, on one side, shows that we can define the semantics of logic operators syntactically, and on the other side, shows the expressiveness of
our approach.

As discussed in the introduction section, our motivation is to
propose a simple, extensible, expressive and efficient KR approach.
While extensibility and expressiveness are discussed in the above
two paragraphs, simplicity is difficult to justify. We argue that
our approach indeed satisfy the three aspects of simplicity. For
primitiveness, our approach only uses naive set theory, syntax
including individuals, concepts and operators and knowledge of the
form $a=b$. For succinctness, the primitive form only needs at most
double length to simulate logic, which is an arguably succinct KR
formalism. For user-friendliness, we believe that knowledge of the
form $a=b$, similar to the assignment statement in programming, can
be easily understood and used by knowledge engineers.

Certainly, one can define multi-assertion, nested assertion and
logic operators directly. Nevertheless, our motivation is to provide
a simple foundation for knowledge representation so that all other
features and building blocks in KR can be defined as extensions of
it. We believe that our primitive form is such a candidate, evident
from the fact that it is able to capture high-order logic
expressions.

This work has two philosophical implications. First, for answering
the question ``what is knowledge", our approach defines it as
equality assertions between two terms. Again, evident from the fact
that single equality assertions can capture high-order logic
expressions based on naive set theory, we believe that it provides a
uniformed view of what knowledge is. Such a uniformed view is
critical for not only understanding and representing knowledge but
also utilizing and reasoning about knowledge. Second, we have shown
that we can define the semantics of logic syntactically based on
naive set theory. We believe that the same thing can be done for the
semantics of other features in KR, e.g., nonmonotonic reasoning.
This is useful as most operations done by machines are syntax based.

This paper is mainly focused on the representation part. We leave
the reasoning part and the efficiency discussions to another paper.
Nevertheless, it is worth mentioning a little here. Roughly
speaking, reasoning is about how to derive properties from a
knowledge base. We distinguish between querying and reasoning. The
former is to check whether an assertion is a property of a knowledge
base, while the latter is to find some properties of a given
knowledge base. Clearly, reasoning can serve as a means
for querying, but they are not the same. Querying is generally
difficult of our approach as it can express logic. Nevertheless,
reasoning could be efficient, and that is exactly the focus of our
reasoning paper.

Although querying the full language is generally undecidable, there
could be some meaningful and useful tractable subclasses. An
important case is database. Note again that, in the primitive form,
knowledge is simply formalized by equality assertions of the form
$a=b$. Nevertheless, this can be indeed expressive as $a$ and $b$
could be complex nested terms. A database under our context only
contains two kinds of equality assertions, i.e., data of the form
$Op(a_1,\dots, a_n)=b$, where $a_i,1 \le i \le n$, and $b$ are
atomic individuals and membership statements $a \in C$, where $a$ is
an atomic individual and $C$ is an atomic concept. In this sense,
data is a special kind of knowledge. Querying on such a database is
tractable. We leave the detailed discussions to another work.

Our set theoretic KR approach is deeply inspired by and rooted in
many other KR formalisms, including propositional and first-order logic, semantic network and description logic. The dynamic
version of this approach (will be presented in another paper) is
deeply related to rule based formalisms including Hoare logic and
answer set programming. Interestingly, although originated from a
different motivation, our approach shares many basic ideas and
borrows many notations from description logic \cite{Baader03}. In
fact, we can rewrite all building blocks in description logic to our
approach since the primitive form can capture first-order logic. Table \ref{dl-sk} depicts some of them.
\vspace{-.15in}
\begin{table}[!hbp]
\caption{Rewriting description logic into our approach}\label{dl-sk}
\begin{center}
\small
\begin{tabular}{|c|c|c|}
\hline Constructs & Description logic & Our approach \\
\hline individual & individual & individual \\
\hline concept & concept & concept  \\
\hline role & Role  & binary Boolean operator \\
\hline
\hline intersection & $C \sqcap D$ & $C \cap D$ \\
\hline union & $C \sqcup D$ & $C \cup D$ \\
\hline complement & $ \neg C$ & $\mathcal{I} \setminus C$ \\
\hline reverse role & $R^-$ & $R^-(C,D) ::= R(D,C)$ \\
\hline existential restriction & $\exists R.C$ & $\mathcal{I}|\widehat{R^-}(\mathcal{I}) \cap C \ne \emptyset$ \\
\hline universal restriction & $\forall R.C$ &
$\mathcal{I}|\widehat{R^-}(\mathcal{I}) \subseteq C $ \\
\hline at least restriction & $\ge n R.C$ & $\mathcal{I}|(\widehat{R^-}(\mathcal{I}) \cap C)^C \ge n$ \\
\hline nominal & $\{a\}$ & $\{a\}$ \\
\hline \hline concept assertion & $C(a)$ & $a \in C$ \\
\hline role assertion & $R(a,b)$ & $R(a,b)$ \\
\hline individual equality & $a \approx b$ & $a=b$ \\
\hline concept inclusion & $C\sqsubseteq D $ & $C \subseteq D$ \\
\hline
\end{tabular}
\end{center}
\end{table}
\vspace{-.1in} Here, $\hat{R}$ is defined to transform a binary
Boolean relationship $R$ to a unary operator, i.e., $\hat{R}(C) ::=
D|R(C,D)$, and $(D)^C$ denotes the cardinality of $D$.

Nevertheless, our approach differs from description logic in several
essential ways. The most important difference is that, for the
purpose of forming new concepts by operators/roles, our approach
directly uses set theoretic constructs including comprehension and
replacement, while description logics use role restrictions. As an
example, suppose that we want to specify a concept including all
human having female children. In our approach, this is formalized by
$Human \D Children(Human) \cap Female \ne \emptyset$, while in
description logic, it is formalized by $\exists Parentof.(Female)$.
Secondly, we use multi-ary operators, e.g., the $Add$ operator,
instead of binary Boolean relationships to connect
individuals/concepts. Thirdly, all knowledge in our approach are
essentially formalized in the same form, i.e., equality assertions.
Fourthly, we allow complex assertions including high-order
constructs. Last but not least, we particularly highlight the
importance of extensibility in our approach.

We shall present a series of papers to propose the set theoretic
knowledge representation approach. This paper is a foundation that is mainly concerned with the
basic ideas and the representation part. As mentioned, there are a
number of things to present in the future. One critical task is to
present the reasoning part. Another one is to formalize
dynamics, including how to represent basic and compound actions, how
to describe the effects of actions and the interactions among
knowledge and actions.

\bibliographystyle{named}
\bibliography{set-kr}

\end{document}